\pdfoutput=1

\documentclass[11pt]{article}

\usepackage{emnlp2021}

\usepackage{times}
\usepackage{latexsym}

\usepackage[T1]{fontenc}

\usepackage[utf8]{inputenc}

\usepackage{microtype}

\usepackage{times}
\usepackage{latexsym}
\usepackage{url}
\usepackage{color,soul}

\usepackage{graphicx}
\usepackage{amsmath}
\usepackage{amssymb}
\usepackage{todonotes}
\usepackage{algorithm}
\usepackage{algorithmic}
\usepackage{flexisym}
\usepackage{makecell}
\usepackage{comment}

\newcommand\nheads{n_{H}}
\newcommand\intermediatedim{d_{I}}
\newcommand\embeddingdim{d_{E}}
\newcommand\Lzero{L_{0}}

\newcommand\ffgates{\Gamma^{\hbox{ff}}}
\newcommand\attngates{\Gamma^{\hbox{attn}}}

\newcommand\lambdaattn{\lambda^{\mathrm{attn}}}
\newcommand\lambdaff{\lambda^{\mathrm{ff}}}

\newcommand\lambdaattnopt{\lambda^{a*}}
\newcommand\lambdaffopt{\lambda^{f*}}

\newcommand\ffp{\mathrm{ff}}
\newcommand\attnp{\mathrm{attn}}

\title{Structured Pruning of BERT-based Question Answering Models}

\author{J.S. McCarley \and Rishav Chakravarti \and Avirup Sil\\
  IBM Research AI\\
  Yorktown Heights, NY\\
  \texttt{\{jsmc,rchakravarti,avi\}@us.ibm.com} \\}

\hypersetup{draft}
\begin{document}
\maketitle
\begin{abstract}
The recent trend in industry-setting Natural Language Processing (NLP) research has been to operate large 
pretrained language models like BERT under strict computational limits. 
While most model compression work has focused on ``distilling" a general-purpose language representation using expensive {\em pretraining distillation},
less attention has been paid to creating smaller task-specific language representations which, arguably, 
are more useful in an industry setting. 
In this paper, we investigate compressing BERT- and RoBERTa-based question answering systems by structured pruning of parameters from the underlying
transformer model. 
We find that an inexpensive combination of  {\em task-specific structured pruning} and {\em task-specific distillation},
without the expense of {\em pretraining distillation},
yields highly-performing models across a range of speed/accuracy tradeoff operating points.
We start from existing full-size models trained for SQuAD 2.0 or Natural Questions and introduce gates that allow selected parts of transformers to be individually eliminated.
Specifically, we investigate (1) structured pruning to reduce the number of parameters in each transformer layer,
(2) applicability to both BERT- and RoBERTa-based models,
(3) applicability to both SQuAD 2.0 and Natural Questions,
and (4) combining structured pruning with distillation.
We achieve a near-doubling of inference speed 
with less than a 0.5 F1-point loss in short answer accuracy on Natural Questions.

\end{abstract}

\section{Introduction}
While knowledge distillation from large pretrained language models (e.g. BERT-large as a teacher)
has mitigated some of the computational burdens of these models, computationally expensive {\em pre-training distillation}
unnecessarily limits the ability of efficient student models to adopt the latest innovations in 
pretrained language models and transformer architecture.
In this paper, we show that a combination of {\em task-specific structured } pruning and {\em task-specific distillation},
yields highly-performing compressed versions of existing models across a range of speed/accuracy tradeoff operating points, 
without the expense of revisiting the pretraining data.


Among Natural Language Processing (NLP) tasks, question answering (QA), in particular, has immediate applications in real-time systems.
A relatively new field in the open domain question answering (QA) community is machine reading comprehension (MRC) which aims to read and comprehend a given text, and then answer questions based on it.
MRC is one of the key steps for natural language understanding. MRC also has wide applications in the domain of conversational agents and customer service support.
Transformer-based models have led to striking gains in accuracy on MRC tasks recently,
as measured on the SQuAD v1.1 \cite{squad} and SQuAD v2.0 \cite{squad2} leaderboards.
We briefly mention three MRC tasks:  SQuAD v1.1 consists of reference passages from Wikipedia with answers and questions constructed
by annotators after viewing the passage.
SQuAD v2.0 augmented the SQuAD v1.1 collection with additional questions that did not have answers in the reference passage.
Natural Questions (NQ) \cite{naturalQuestions} removed the observational bias by starting from questions submitted to Google and providing annotated answers from appropriate passages. 

\begin{table*}
\begin{center}
\begin{tabular}{c|c|c|c|c}
\hline
\textbf{model} &  \textbf{params} & \textbf{SQuAD 1.1} & \textbf{SQuAD 2.0} & \textbf{NQ}\\
\hline
 BERT-large & 356M & 90.9 (c)  & 83.52 & 56.14 \\ 
  BERT-base & 125M & 88.4 (a) 88.5 (b)  & 76.4 (a) & 52.75\\
  DistilBert & 63M & 86.2 (a) 86.9 (b) & 69.5 (a) & 50.46 \\
 TinyBert  & 63M & 87.5 (a) & 73.4 (a) & 44.64 \\
 \hline
\end{tabular}
\end{center}
\caption{Comparison of published F1 scores of well-known distillation's of BERT on several question-answering tasks. Though not strictly comparable,
we observe that on SQuAD 1.1 smaller models approach BERT-large in accuracy, whereas the smaller models underperform notably on both SQuAD 2.0 and 
Natural Questions (NQ). We show the short answer F1 for NQ.
Sources: (a)=\cite{jiao2019tinybert}, (b)=\cite{sanh2019distilbert}, (c)=\cite{bert}.}
\label{qaIsHard}
\end{table*}

MRC seems to be a particularly difficult task to speed up.
While distillation papers have advertised impressive speedups with near-negligible loss in accuracy on GLUE benchmarks, published applications of 
distillation to MRC have been less impressive (often relegated to the appendix.)
In Table \ref{qaIsHard}, we compare the accuracies (F1 score) of Distilbert and TinyBert, two well-known compressions of BERT, with baseline ("out-of-the-box" pretraining) BERT-large and BERT-base models on three MRC tasks, 
using the number of parameters as a crude proxy for speed. 
\footnote{MobileBert \cite{sun-etal-2020-mobilebert} required extensive pretraining architecture search experiments in order to customize the
teacher model, and does not represent a fair comparison when the goal is to compress {\em existing} models.}
Compared to BERT-large, models with fewer parameters achieved modest losses on SQuAD 1.1.
The shortfalls on the more challenging SQuAD 2.0 were much larger.  We also note that the shortfalls of smaller models were large on NQ.
SQuAD is also seen as a worst-case performance loss for speed up techniques based on quantization, \cite{qbert} 
while the difficulty of distilling a SQuAD model (compared to sentence-level GLUE tasks) is acknowledged in \cite{jiao2019tinybert}.
We speculate that these difficulties are because answer selection via pointer networks requires token level predictions
rather than passage level classification,
and requires long range attention between query and passage.


%
%



The contributions of this paper are
\begin{enumerate}
    \item Application of structured pruning techniques to the hidden dimension of the feed-forward layer, not just the attention heads \cite{sixteenHeads},
\item the combination of distillation and pruning,
\item thereby significantly pruning the
MRC system with minimal loss of accuracy and considerable speedup, all without the expense of revisiting pretraining \cite{sanh2019distilbert,jiao2019tinybert}
\end{enumerate}

\noindent Furthermore we survey multiple pruning techniques (both heuristic and trainable)
and provide recommendations specific to transformer-based question answering models.
We focus exclusively on structured pruning \cite{structuredConv} to avoid sparsity issues.
During the course of the investigation, we also learn that 
an optimal pruning learns a structure
consisting of {\em non-identical} transformers,
namely lightweight transformers near the
top and bottom while retaining more complexity in the intermediate layers,
instead of the
typically 12-24 layers of {\em identically}
sized transformers, common in widely distributed pre-trained models

\section{Related work}

The field of neural networks compression has been extensively reviewed in \cite{neill2020overview}.  Here we focus on results relevant to MRC.
While distillation (student-teacher) of BERT has produced notably smaller and faster models
\cite{tang2019distilling,turc2019wellread,tsai2019small,yang2019model},
the focus has been on passage level annotation tasks (e.g. GLUE) that do not require long-range attention links between query and passage.

Distillation of typical MRC models has been much more limited:
DistilBERT \cite{sanh2019distilbert} used pretraining distillation to obtain $60\%$ speedups on GLUE tasks while retaining $97\%$ 
of the accuracy.  However, MRC results, after additional task-specific distillation, included a modest speedup and small performance loss on SQuAD 1.1.
TinyBERT \cite{jiao2019tinybert} used both pretraining and task-specific distillation to obtain $9.4 \times$ speedups on GLUE.
However, they restricted SQuAD evaluation to using BERT-base as a teacher, and deferred deeper investigation to future work.
MobileBERT \cite{sun-etal-2020-mobilebert} obtains strong results after an extensive architecture search in order to construct
a teacher model with custom architecture which is both pre-trained and used
for pretraining distillation of the student model.  This approach 
represents a notable increase in pretraining expense, and is further removed from this paper's goal of shrinking {\em existing} models.
\cite{turc2019wellread} investigated pretraining and distilling smaller models from scratch, but tested only on passage-level annotation tasks.
The authors are not aware of any results from distilled models on NQ.








Investigations into pruning BERT have also omitted MRC.
Michel et al.~\shortcite{sixteenHeads} applied simple gating
heuristics to prune BERT attention heads and achieve speedups
on MT and MNLI.
Voita et al.~\shortcite{voita-etal-2019-analyzing}
introduced $L_{0}$ regularization to BERT while focusing on linguistic interpretability of attention heads but 
did not report speedups.
$L_0$ regularization was combined with matrix factorization to prune transformers for classification in \cite{wang2019structured}.
Gale et al.~\shortcite{state_of_sparsity} induced unstructured sparsity on a transformer-based MT model,
but did not report speedups.
Kovaleva et al.~\shortcite{darkSecretsOfBert} also focused on interpreting attention, 
and achieved small accuracy gains on GLUE tasks by disabling (but not pruning) certain attention heads.
Structured pruning as a form of dropout is explored in  \cite{fan2020reducing}.
They prune entire layers of BERT, but suggest that smaller structures could also be pruned. 
They evaluate on MT, language modeling, and generation-like tasks, but not SQuAD.

Another set of approaches omit cross-attention between documents and queries in the lower layers so that precomputed document representations can be used at inference time.  These approaches report results only on SQuAD 1.1
\cite{cao-etal-2020-deformer}
and various IR tasks
\cite{10.1145/3397271.3401075,MacAvaney_2020}
, but not SQuAD 2.0 or NQ.

Other approaches to speeding up transformers include ALBERT \cite{albert}, which shared parameters across layers in order to accelerate training, but did not report timings of inference.


QBERT \cite{qbert} and Q8BERT \cite{zafrir2019q8bert} aggressively quantized floating point calculations to ultra-low precision in order to compress BERT.
They noted that SQuAD was harder to quantize (greater performance drop) than other tasks.

Finally, \cite{li2020train} investigated both unstructured pruning and quantization of RoBERTa as a function of
model size, and found that both pruning and quantization were complementary, an important reminder that multiple 
types of compression are not mutually exclusive.
Very recently, \cite{kim-hassan-2020-fastformers} combined distillation, structured pruning, and quantization and achieved impressive speedup on both CPU and GPU on GLUE tasks, but did not report results on SQuAD/NQ-style question answering.

\section{Pruning transformers}


\subsection{Gate placement}
Our approach to pruning consists of inserting additional {\em trainable} parameters, {\em masks}, into a transformer.  The value of each mask variable controls whether an entire block of transformer parameters (e.g. an attention head) is used by the model.
Specifically, each mask is a vector of gate variables $\gamma_i \in [0,1]$, pointwise multiplied into a slice of transformer parameters, where $\gamma_i=1$ allows a slice to remain active,
and $\gamma_i=0$ deactivates the slice.
We insert two types of masks into each transformer.
We describe the placement of each mask with the terminology of \cite{vaswani2017attention},
indicating relevant sections of that paper.

In each self-attention sublayer, we place a mask, $\attngates$ of size $\nheads$ which selects attention heads to remain active. (section 3.2.2)

In each feed-forward sublayer, we place a mask, $\ffgates$ of size $\intermediatedim$ which selects ReLU/GeLU activations to remain active. (section 3.3)


Here $\nheads$ is the number of heads per transformer layer (12 or 16), $\embeddingdim$ is the the size of the embeddings (768 or 1024) as well as  the inner hidden dimension, and 
$\intermediatedim$ is the size of the intermediate activations in the feed-forward part of the transformer (3072 or 4096.)  Sizes are for (BERT-base, BERT-large).

\begin{algorithm*}[!htb]
\small
\caption{Pruning an $L_0$ regularized model: $\mathrm{ff}(\hbox{Sq}) + \mathrm{attn}(\hbox{Sq}) + \mathrm{retrain(Sq)}$}
\begin{algorithmic}[1]
\REQUIRE $\langle BERT_{QA}, D, \lambdaattn, \lambdaff$ \\
\COMMENT{ $BERT_{QA}$ is an already-trained BERT question answering model that will be pruned, $D$ is question-answering (SQuAD) training data, $\lambdaattn$ and $\lambdaff$ are penalty weights that determine how much to prune  }
\STATE $\alpha^{\mathrm{attn}}_i \leftarrow \langle BERT_{QA}, D \rangle 
\hspace*{\fill} \rhd$ train attention gate parameters by optimizing  $\mathcal{L} + attn(\lambdaattn)$ \\
\STATE $\alpha^{\mathrm{ff}}_i \leftarrow \langle BERT_{QA}, D \rangle 
\hspace*{\fill} \rhd$ train feed-forward gate parameters by optimizing $\mathcal{L} + ff(\lambdaff)$ \\
\STATE $\attngates \leftarrow threshold(\alpha^{\mathrm{attn}})
\hspace*{\fill} \rhd$ select final gate values for attention heads\\
\STATE $\ffgates \leftarrow threshold(\alpha^{\mathrm{ff}})
\hspace*{\fill} \rhd$ select final gate values for feed forward heads\\
\STATE $BERT_{QA}\textprime  \leftarrow \langle BERT_{QA}, \attngates \rangle
\hspace*{\fill} \rhd$ prune the attention heads\\
\STATE $BERT_{QA}\textprime{}\textprime{}  \leftarrow \langle BERT_{QA}\textprime, \ffgates \rangle
\hspace*{\fill} \rhd$ prune the feedforward layers\\
\STATE $BERT_{QA}\textprime{}\textprime{}\textprime{}  \leftarrow \langle BERT_{QA}\textprime{}\textprime{}, D \rangle
\hspace*{\fill} \rhd$ continued training of remaining BERT parameters subject to $\mathcal{L}$ \\

\end{algorithmic}
\label{pruning-algorithm}
\end{algorithm*}

\subsection{Determining Gate Values}
\label{ssec:gate-values}
We investigate four approches to determining the gate values.

(1) Random: each $\gamma_i$ is sampled from a Bernoulli distribution of parameter $\alpha$,
where $\alpha$ is manually adjusted to control the sparsity.  This method is the naive baseline,
and is expected to be worse than other methods.

(2) Gain:  We follow the method of \cite{sixteenHeads} and estimate the influence of each gate $\gamma_i$
on the training set likelihood $\mathcal{L}$ by treating each $\gamma_i$ as a continuous parameter and
computing the mean  
\begin{equation}
g_i={\left |{ \frac{\partial \mathcal{L}}{\partial \gamma_i}} \right | _ {\gamma_i=1}}
\end{equation}
(``head importance score'') during one pass over the training data.
We threshold $g_i$ to determine which transformer slices to retain.


(4) $L_0$ regularization:  Following the method described in  \cite{louizos2018learning},
the gate variables $\gamma_i$  are sampled
\begin{equation}
    \gamma_i \sim \mathrm{hc}(\alpha_i)
\end{equation}
from a hard-concrete distribution $\hbox{hc}(\alpha_i)$
\cite{concrete} parameterized by a corresponding variable $\alpha_i \in \mathbb{R}$.
The $\alpha_i$ are trained by optimizing
the task-specific objective function $\mathcal{L}$ (typically cross-enropy) penalized in proportion to the number of expected instances of $\gamma=1$, with
proportionality constants  $\lambdaattn$ in the penalty terms $\mathrm{attn}(\lambdaattn)$, e.g.
\begin{align}
\begin{split}
\mathcal{L}_{\mathrm{cross\mbox{-}entropy}} & +
    \mathrm{attn}(\lambdaattn)  = \mathcal{L}_{\mathrm{cross\mbox{-}entropy}} \\
    & -
    \lambdaattn \mathbb{E}_{\mathrm{hc}}\left[ \sum_{i} \delta_{\gamma_i, 1} \right]
    \label{eq:objPenalty}
\end{split}
\end{align}
(and similarly for the $\mathrm{ff}(\lambdaff)$.) The $\lambda$ are hyperparameters  controlling the sparsity.
The expectation is over the same hard-concrete distribution from which we sample.
We resample the $\gamma_i$ with each minibatch.
This objective function is differentiable with respect to the $\alpha_i$
because of the reparameterization trick. \cite{kingma2013autoencoding,pmlr-v32-rezende14}
The $\alpha_i$ are updated by backpropagation for up to one training epoch on the task training data,
with all other transformer parameters held fixed.  The final values for the gates $\gamma_i$ are obtained by
thresholding the $\alpha_i$.
We note that either the log-likelihood or a distillation-based objective can be penalized
as in Eq. (\ref{eq:objPenalty}).
The cost of training the gate parameters is comparable to extending fine tuning for an additional epoch.

\subsection{Structured Pruning}
After the values of the $\gamma_i$ have been determined by one of the above methods, we prune the model.
Attention heads corresponding to $\gamma^{\mathrm{attn}}_{i}=0$ are removed.
Slices of both linear transformations in the feed-forward sublayer which correspond to $\gamma^{\mathrm{ff}}_i=0$ are
removed.
The pruned model no longer needs masks, and now consists of smaller transformers of varying,
{\em non-identical} sizes.  For experiments on some hardware, matrices are forced to have sizes that are round numbers
rather than strictly respecting the threshold.

\subsection{Extended training}
As noted by \cite{structuredConv}, the task-specific training of all parameters of a pruned model may be continued further with the (unpenalized) task-specific objective function $\mathcal{L}.$
In some experiments we continue training by incorporating distillation: the unpruned model
is the teacher, and the pruned model is the student.

In summary, the entire pruning procedure, starting from a trained model for an MRC task, consists of
\begin{enumerate}
\item{ inserting masks into each transformer layer}
\item{ determining values of the masks, either heuristically (methods (1)-(3)) or training them with penalized objective functions (method (4))}
\item{ replacing transformer parameter matrices with smaller matrices, pruned according the masks determined in the previous step}
\item{Either}
\begin{itemize}
\item{ continued training of the pruned transformer parameters with the original objective function ($retrain$)}
\item{ {\em or} continued training with a distillation objective function ($distill$), using the original unpruned model as the teacher}
\end{itemize}
\end{enumerate}

This algorithm is presented in pseudocode in Algorithm \ref{pruning-algorithm}.

\section{Experiments}
\subsection{Overall Setup and outline}
We evaluate our proposed method on two benchmark QA datasets: SQuAD 2.0 \cite{squad2} and Natural Questions (NQ) \cite{naturalQuestions}. SQuAD 2.0 is a dataset of questions from Wikipedia passages, proposed by human annotators while viewing these Wikipedia passages. NQ is a dataset of Google search queries with answers from Wikipedia pages provided by human annotators. Of the two, NQ is more natural, as the questions were asked by humans on Google without having seen the passage. On the other hand, SQuAD  annotators read the Wikipedia passage first and then formulated the questions. 


We address several empirical questions here: 1. Do techniques developed on BERT-base transfer to BERT-large?  2. Do the proposed techniques transfer across datasets? 3. Does incorporating a distillation objective further improve our model's performance?

To answer these we tune our hyper-parameters on a subset of SQuAD 2.0 using a BERT-base model,
and then test them on the full SQuAD 2.0 with a BERT-large model. Further, we show that the same techniques are applicable on the NQ dataset not just with BERT but also with RoBERTa. Finally we show that incorporating distillation achieves even stronger and more flexible results. When practical we report numbers as an average of 5 seeds.

\begin{table*}[t]
\small
\begin{center}
\begin{tabular}{|lrr||rrrrrr|}
\hline
\thead{pruning of \\ BERT-large} & 
\thead{$\frac{\lambdaattn}{\lambdaattnopt}$} &
\thead{$\frac{\lambdaff}{\lambdaffopt} $} &
\thead{time\\sec.} &
\thead{F1\\$+$retrain} &
\thead{F1\\no retrain} &
\thead{\% attn\\ removed} &
\thead{\%ff\\ removed} &
\thead{size\\(MiB)} \\
\hline
$a$: no pruning & 0 & 0 & 2712 & & 84.6 & 0 & 0 & 1279 \\
$b: \attnp(\hbox{Sq})$ & 1 & 0 & 2288 & & 84.2 & 44.3 & 0 & 1112 \\
$c: \ffp(\hbox{Sq})$ & 0 & 1 & 2103 & & 83.2 & 0 & 48.1 & 908 \\
\hline
$d: \ffp(\hbox{Sq}) + \attnp(\hbox{Sq}) $ & 1 & 1 & 1667 & 83.7 & 82.6 & 44.0 & 48.1 & 740 \\
$e: \ffp(\hbox{Sq}) + \attnp(\hbox{Sq}) $ & 2 & 2 & 1391 & 83.2 & 80.9 & 53.1 & 64.9 & 576 \\
$f: \ffp(\hbox{Sq}) + \attnp(\hbox{Sq}) $ & 3 & 3 & 1213 & 82.4 & 76.8 & 57.6 & 73.7 & 492 \\
$g: \ffp(\hbox{Sq}) + \attnp(\hbox{Sq}) $ & 4 & 4 &1128 & 81.5 & 67.8 & 60.1 & 78.4 & 441 \\
\hline
\end{tabular}
\end{center}
\caption{Decoding times, accuracies on SQuAD 2.0, and space savings achieved at sample operating points of pruned {\em BERT large-qa}, with and without continued training. }
\label{large-model-table}
\end{table*}


\subsection{SQuAD 2.0}
\subsubsection{Experimental Setup and hyper-parameters}
For selection of hyper-parameters (learning rate and penalty weight exploration) and in order to minimize overuse of the official dev-set, 
we use $90\%$ of the official SQuAD 2.0 training data for training gates,
and report results on the remaining $10\%$. This resulting model ({\em base-qa}) is initialized from a {\em bert-base-uncased} SQuAD 2.0 system trained on the $90\%$ with a baseline performance of F1 = 75.0 on the $10\%$ dataset.
Experiments described were implemented using code from the HuggingFace repository \cite{Wolf2019HuggingFacesTS} and incorporated either {\em bert-base-uncased} or {\em bert-large-uncased} with a standard task-specific head.

Our final SQuAD 2.0 model ({\em large-qa}) use the standard train/dev split of SQuAD 2.0 and is initialized from a {\em bert-large-uncased} system trained with the method described in \cite{glass2019span}. It achieves an F1 = 84.6 on the official dev set, somewhat exceeding "out-of-the-box" BERT question answering models.



The gate parameters of the $L_0$ regularization experiments are trained for one epoch
starting from the models above, with all transformer and embedding parameters fixed.
We investigated learning rates of $10^{-3}$, $10^{-2}$,
and $10^{-1}$ on {\em base-qa}, and chose $10^{-1}$ for presentation and results on {\em large-qa}.
This is notably larger than typical learning rates to tune BERT parameters.
We used a minibatch size of 24 and otherwise default hyperparameters of the BERT-Adam optimizer.
We used identical parameters for our {\em large-qa} experiments, except with gradient accumulation of 3 steps.




\subsubsection{Accuracy as function of pruning}

\begin{figure}
  \centering
  \includegraphics[width=1\linewidth]{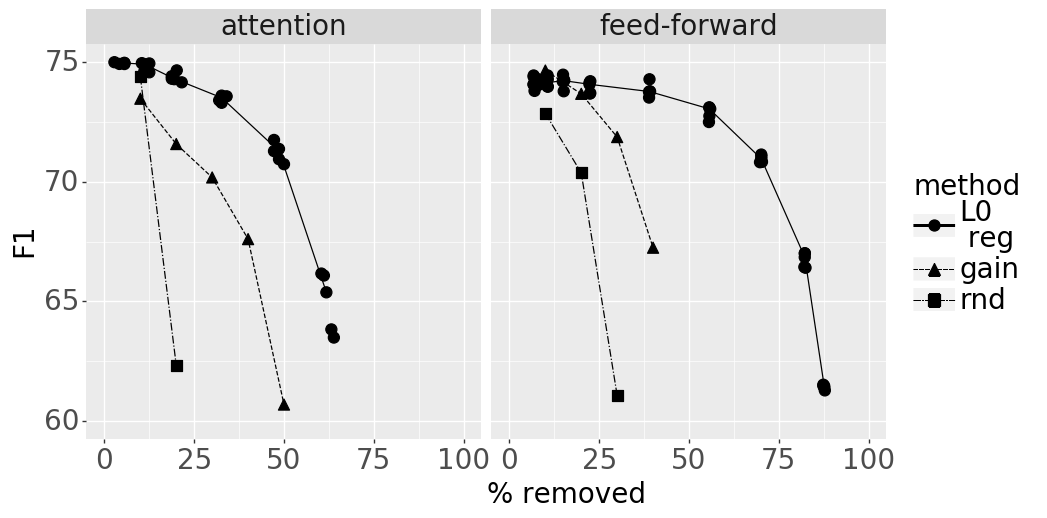}
   \caption{Comparison of pruning methods on SQuAD 2.0: F1 vs percentage of attention heads and feed forward activations pruned from {\em base-qa}}
   \label{ff_joint_pct}
\end{figure}

In Figure \ref{ff_joint_pct} we plot the {\em base-qa} F1 as a function of the percentage of heads removed. The performance of `random` decays 
abruptly. 
'Gain' is better.
$\Lzero$regularization is best, allowing $48\%$ pruning at a cost under $5$ F1-points.

Also in Figure \ref{ff_joint_pct} we plot the (accuracy) $F1$ of removing feed-forward activations.  We see broadly similar trends as above, except that the performance is robust to even larger pruning. 
As before $\Lzero$ regularization is best, allowing $70\%$ pruning at cost under $5$ F1-points.



\subsubsection{Validating these results}

On the basis of the development experiments, we select an operating point, namely the largest values of $\lambdaattn$
and $\lambdaff$ with $<5$ F1-point loss.  
After rescaling to the larger model size, we denote the weights
as $\lambdaattnopt=1.875 \times 10^{-3}$ and $\lambdaffopt=7.5\times 10^{-6}.$
We train the feed-forward and attention gates of {\em large-qa} with these penalties,
as well as multiples $2\times$, $3\times$, and $4\times$.
The decoding times, accuracies, and model sizes are summarized in Table \ref{large-model-table}.
Accuracies are medians of 5 seeds, and timings are medians of
5 decoding runs with the median seed, on a single Nvidia K80 with
batch size 1.
Models in which both attention and feed-forward components are pruned were built from the
{\em independently trained} gate configurations of attention and feed forward.
For corresponding penalty weights, the {\em large-qa} was pruned somewhat less than {\em base-qa},
and the $F1$ loss due to pruning was smaller.

Much of the loss in accuracy is recovered by continuing the training for an additional epoch (column 5) after the pruning, even though the accuracy without retraining (column 6) decreases
substantially
as more is pruned.  We highlight the operating point 
of Table \ref{large-model-table}, row $e$ , which after continued training,
loses less than 1.5 $F1$ points, while nearly doubling the decoding speed.


\subsubsection{ Impact of pruning each layer}
\begin{figure}[t]
\begin{center}
\includegraphics[width=1\linewidth]{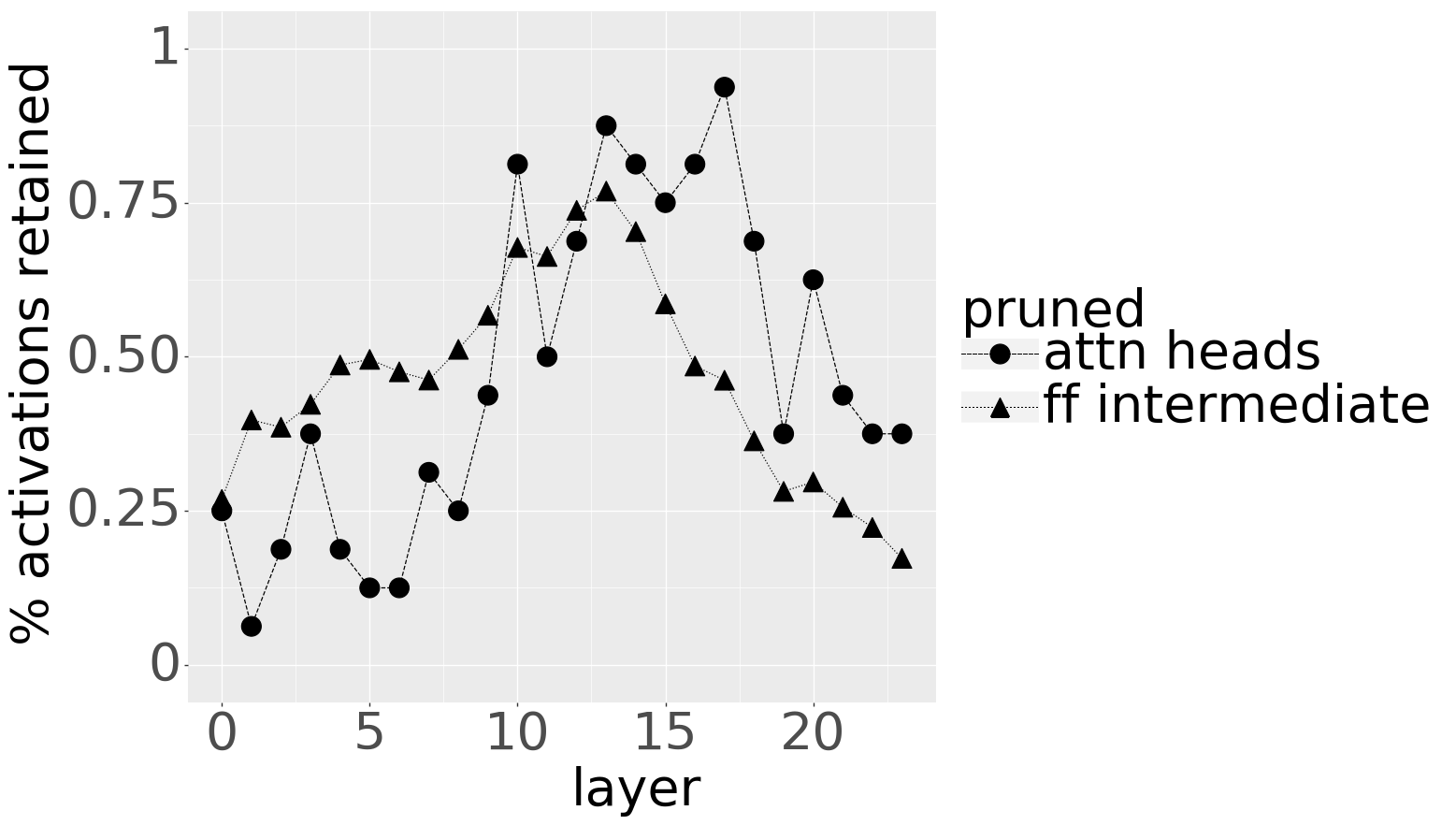}
\end{center}
\caption{Percentage of attention heads and feed forward activations remaining after pruning, by layer}
\label{what_was_pruned}
\end{figure}

In Fig. \ref{what_was_pruned} we show the percentage of attention heads and feed forward activations remaining after pruning, by layer.
We see that intermediate layers retained more, while layers close to the embedding and close to the answer were pruned more heavily.

\subsection{Natural Questions}


We address three questions in this section:

(1) Are the pruning techniques developed for the SQuAD 2.0 task also applicable to the NQ task?

(2) Do pruning techniques developed for BERT also apply to RoBERTa?

(3) Can we combine distillation and pruning to achieve even smaller, faster models?

\subsubsection{Transfer of gates}

We take the pruned BERT-large models described above and use the identical model parameters
as the initialization for continued training (using the cross-entropy objective function)
of an NQ model.  In other words, the gate variables are trained on SQuAD 2.0, and the 
only use of the NQ training data is in the continued training of the remaining transformer parameters,
denoted $retrain(NQ)$.
The results shown in Table \ref{prunedBERTcontNQ}, while far from optimal, are encouraging.
They suggest that the redundancies in BERT that are removed by pruning are not task-specific or domain-specific and
that a pruned model is relatively robust.


\subsubsection{RoBERTa}

\begin{table*}
\small
\begin{center}
\begin{tabular}{|lrr||r|r|r|r|}
\hline
\thead{pruning of \\ BERT-large} & 
\thead{$\frac{\lambdaattn}{\lambdaattnopt}$} &
\thead{$\frac{\lambdaff}{\lambdaffopt} $} &
\thead{\% attn\\ removed} &
\thead{\%ff\\ removed} &
\thead{LA\\(F1)} &
\thead{SA\\(F1)} \\
\hline
$ \hbox{no pruning}    $ & 0 & 0 & 0 & 0 & 66.1 & 54.7 \\    
$ \ffp(\mathrm{Sq}) + \attnp(\mathrm{Sq}) + \mathrm{retrain}(\mathrm{NQ}) $ & 2 & 2 & 44 & 48 & 65.9 & 51.7 \\  
$ \ffp(\mathrm{Sq}) + \attnp(\mathrm{Sq}) + \mathrm{retrain}(\mathrm{NQ}) $ & 4 & 4 &53 & 65 & 64.2 & 49.6 \\  
\hline
\end{tabular}
\end{center}
\caption{NQ accuracy of BERT models pruned on SQuAD, continued cross-entropy training on NQ}
\label{prunedBERTcontNQ}
\end{table*}

RoBERTa-based models have achieved notably higher accuracy than BERT-based models
across a variety of tasks \cite{RoBERTa}, including MRC.
For example, on NQ short answers, our RoBERTa-large model achieves $58.8$ - over $4$ F1-points better than the comparable
BERT-large model, which achieved $54.7$.
RoBERTa has the same topology as BERT.  It differs slightly in such aspects as 
tokenization, training data (during pretraining) and training procedure.  The nature
of these differences suggests that the pruning techniques developed for BERT 
should continue to work largely unchanged with RoBERTa.   However, as noted by \cite{RoBERTa},
BERT is significantly undertrained, which raises the concern that RoBERTa might achieve
its better performance by more effectively utilizing the transformer parameters 
that were under-utilized and prunable in BERT.

We pruned this RoBERTa-large NQ model, using the same techniques as described above,
selecting the gate values by $L_0$ regularization for one epoch on approximately $20\%$ of 
the NQ training data, and continued training for an epoch on the full NQ training set.
In Table \ref{prunedRoBERTacontNQ} we show the accuracy and the amount pruned.
We found that to have a similar percentage of parameters pruned, we needed smaller values of $\lambdaattn$ and $\lambdaff$ when training the pruning on NQ, compared to training the pruning on SQuAD.
The loss in accuracy for comparable amounts of pruning is similar to that observed in BERT/SQuAD experiments, indicating that RoBERTa models can be pruned successfully with these techniques.


\begin{table*}
\small
\begin{center}
\begin{tabular}{|lrr||r|r|r|r|}
\hline
\thead{pruning of \\ BERT-large} & 
\thead{$\frac{\lambdaattn}{4\lambdaattnopt}$} &
\thead{$\frac{\lambdaff}{15\lambdaffopt} $} &
\thead{\% attn\\ removed} &
\thead{\%ff\\ removed} &
\thead{LA\\(F1)} &
\thead{SA\\(F1)} \\
\hline
$ a: \hbox{no pruning}    $ & 0 & 0 & 0    & 0    & 70.3 & 58.8 \\
$ b: \ffp(\hbox{NQ}) + \attnp(\hbox{NQ}) + \mathrm{retrain}(\hbox{NQ})    $ & 1 & 2 & 42 & 40 & 68.3 & 57.7 \\
$ c: \ffp(\hbox{NQ}) + \attnp(\hbox{NQ}) + \mathrm{retrain}(\hbox{NQ})    $ & 2 & 4 & 53 & 56 & 67.8 & 55.5 \\
$ d: \ffp(\hbox{NQ}) + \attnp(\hbox{NQ}) + \mathrm{retrain}(\hbox{NQ})    $ &4 & 10 & 68 & 75 & 65.2 & 52.2 \\
\hline
\end{tabular}
\end{center}
\caption{NQ accuracy of RoBERTa models pruned on NQ, continued cross-entropy training on NQ}
\label{prunedRoBERTacontNQ}
\end{table*}


\subsubsection{Combining distillation and pruning}

The simplest way to combine distillation with pruning is, after the model has been pruned, to replace the continued training ($retrain(NQ)$)
by continued training ($distill(NQ)$) with a distillation objective.
Here the unpruned model acts as the teacher and the pruned model is the student.
In Table \ref{prunedRoBERTacontNQdist}, we show results using distillation only in the continued training phase.  
Line $c$ is especially notable - a $2.9$ F1-point gain compared to line $c$ in Table \ref{prunedRoBERTacontNQ}, with less than $0.5$ F1-point loss relative to unpruned, while approaching a doubling of speed.
Timings are median of 5 decoding runs over the entire NQ developement set on an NVidia V100 using 16-bit floating point with batch size 64.
In this experiment, matrices were forced to have sizes that are round numbers, resulting in small changes ($<1\%$) in reported pruning fractions.
We also include for comparison RoBERTa-base model (line $e$) that has been similarly distilled using
RoBERTa-large as a teacher.


\begin{table*}
\small
\begin{center}
\begin{tabular}{|lrr||r|r|r|r|r|}
\hline
\thead{pruning of \\ BERT-large} & 
\thead{$\frac{\lambdaattn}{4\lambdaattnopt}$} &
\thead{$\frac{\lambdaff}{15\lambdaffopt} $} &
\thead{\% attn\\ removed} &
\thead{\%ff\\ removed} &
\thead{LA\\(F1)} &
\thead{SA\\(F1)} &
\thead{time\\sec.}\\
\hline
$ a:\hbox{no pruning}    $ & 0  & 0 & 0 0    & 0    & 70.3 & 58.8 & 2789\\
$ b: \ffp(\hbox{NQ}) + \attnp(\hbox{NQ}) + \mathrm{distill}(\hbox{NQ})       $ & 1 & 2 & 42 & 40     & 69.8 & 58.4 & 1867\\
$ c: \ffp(\hbox{NQ}) + \attnp(\hbox{NQ}) + \mathrm{distill}(\hbox{NQ})       $ & 2 & 4& 53 & 55     & 69.3 & 58.4 & 1523\\
$ d: \ffp(\hbox{NQ}) + \attnp(\hbox{NQ}) + \mathrm{distill}(\hbox{NQ})       $ & 4 & 10 & 68 & 75     & 67.6 & 55.4 & 1135\\
\hline
$ e: \hbox{RoBERTa-base}        $ & NA & NA & NA   & NA   & 67.3 & 55.9 & 1151\\
\hline
\end{tabular}
\end{center}
\caption{RoBERTa models pruned on NQ, continued training on NQ by distillation from unpruned model}
\label{prunedRoBERTacontNQdist}
\end{table*}


Alternately, the pruning phase itself may be driven by a distillation objective.
Here we replace the cross-entropy term in Eq.(\ref{eq:objPenalty}) with a distillation objective
function, and prune the model based on the modified objective function.
We will denote distillation-driven pruning $prune(distillation)$, in contrast to $prune(cross\mbox{-}entropy)$.
All experiments with $prune(distillation)$ involved distillation continued training $distill(NQ).$
When the pruning phase itself is driven by the distillation objective, the results
are not directly comparable because the same values of $\lambdaattn$ 
and $\lambdaff$ yield significantly less pruning for $prune(distillation)$ than for $prune(cross\mbox{-}entropy)$.

\begin{figure}[t]
\begin{center}
\includegraphics[width=1\linewidth]{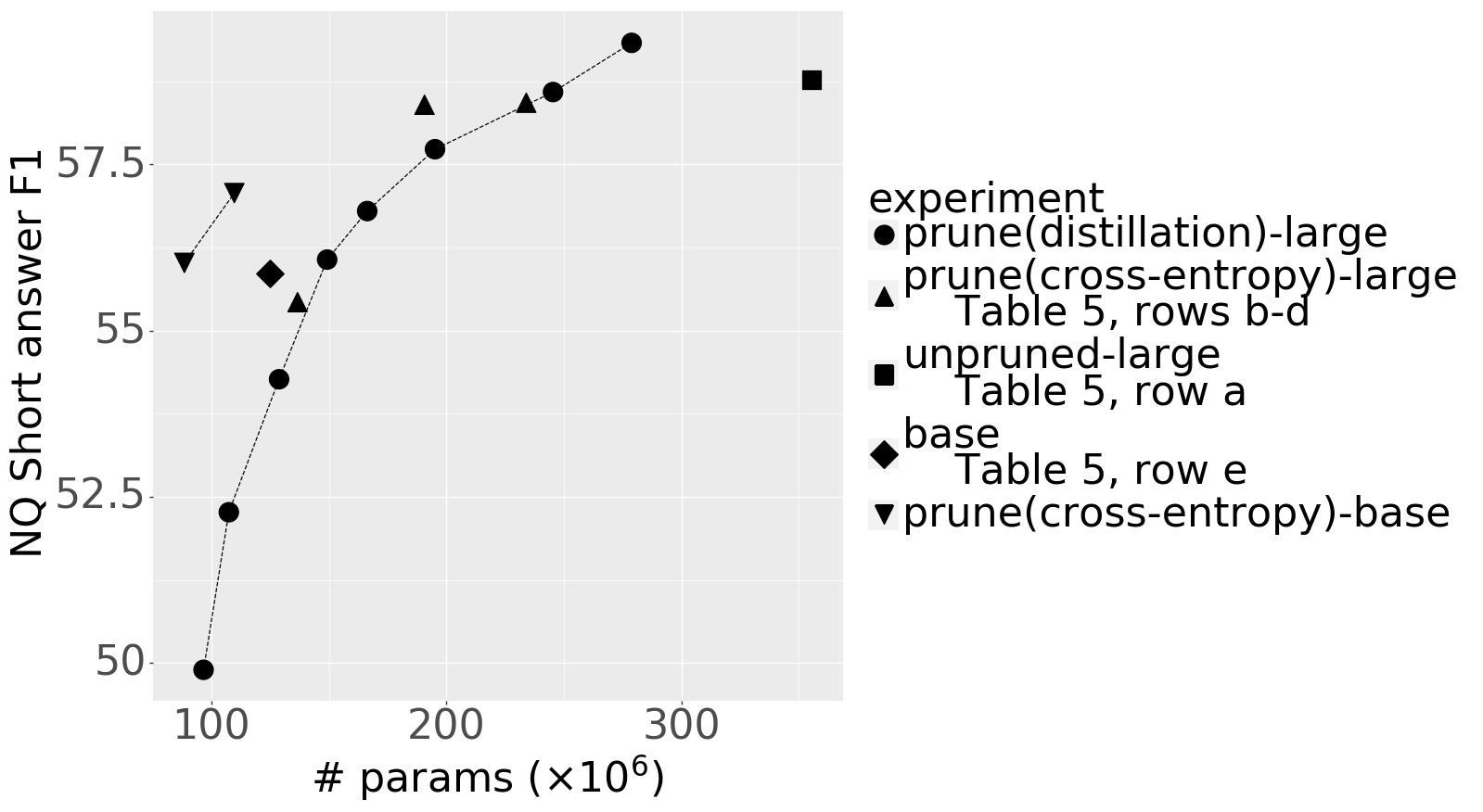}
\end{center}
\caption{Short answer accuracy vs number of parameters (millions), contrasting distillation-driven pruning with cross-entropy-driven pruning. (See text.)}
\label{sa_vs_mparams}
\end{figure}

In Fig. \ref{sa_vs_mparams} we plot the performance of various pruned models as a function of the number of remaining parameters. 
(We have found that the number of parameters is well-correlated with the decoding time for this range of parameters.)
The points labeled {\em prune(distillation)-large} represent various degrees of distillation pruning followed by $distill(NQ)$ continued training of a RoBERTa-large model.
The points labeled {\em prune(cross\mbox{-}entropy)-large} represent various degrees of cross-entropy pruning followed by $distill(NQ)$ (corresponding to Table \ref{prunedRoBERTacontNQdist}, rows $b\mbox{-}d$) of the same
initial model.
The unpruned RoBERTa-large model of Table \ref{prunedRoBERTacontNQdist}, row $a$ is the point {\em unpruned-large} at the far right of the graph.
The distillation pruning does not provide a notable improvement over the cross-entropy driven pruning,
unlike the case of distillation-driven continued training vs cross-entropy driven continued training.

The point labeled {\em base} in Fig.\ref{sa_vs_mparams} is the RoBERTa-base model (line $e$ of Table \ref{prunedRoBERTacontNQdist}) trained with the same distillation technique as our pruned models.
It lies above and to the left
of the envelope of {\em large prune(distillation)} points, which suggests that the pruning+distillation processes are not quite achieving
full potential. 
On the other hand, the pruning$+$distillation processes offer more flexibility of operating
points, without requiring expensive masked language model pretraining at each size.

The pruning-distillation process may also be applied to the RoBERTa-base model {\em base+dist}, and this is illustrated
by the points {\em prune(cross-entropy)-base} in Fig \ref{sa_vs_mparams}.  These points lie even further above and to the left of the
envelope of {\em base} points, and point the way to even smaller and faster NQ models achievable by a combination 
of distillation and structured pruning.
For comparison, the results we have obtained for DistilBERT ($50.46$) and TinyBERT ($44.64$) are at or below 
the bottom edge of this graph.

Averaging across five different initializations (random seeds) of gate parameters, 
a sample operating point for {\em prune(distillation)-large} has attention heads pruned by $60.0 \pm 1.0\%$, feed-forward activations pruned by
$71.9 \pm 0.1\%$ yielding long-answer (LA) F1 of $68.2\pm0.2\%$ and short-answer (SA) F1 of $56.2\pm0.2\%$.
Similarly, a sample operating point for {\em prune(cross-entropy)-base} has attention heads pruned by $20.3\pm1.7\%$, feed-forward
activations pruned by $17.7\pm0.4\%$, yielding long-answer (LA) F1 of $68.0\pm0.2\%$ and short-answer (SA) F1 of $57.0\pm0.2\%$.



\section{Conclusions}
We investigate various methods to prune existing transformer-based MRC models, and evaluate the accuracy-speed tradeoff
for these prunings.
We find that both the attention head layers and especially the feed forward layers can be pruned considerably
with minimal lost of accuracy.

We find that $L_0$ regularization pruning is particularly effective for pruning these two transformer components,
compared to the more heuristic 'Gain' method.
The pruned feed-forward layer and the pruned attention heads are easily combined.
Especially after retraining, this combination yields a considerably faster question answering model with minimal loss
in accuracy.
One operating point nearly doubles the decoding speed on SQuAD 2.0, with a loss of less than $1.5$ F1-points.

The same methods that worked with a BERT-based SQuAD 2.0 model also yield strong results when applied to 
a RoBERTa-based NQ model.  The best performance is achieved by combining distillation with structured pruning.
One operating point almost doubles
the inference speed of RoBERTa-large based model for Natural Questions, while losing less than 0.5 F1-point on short answers, less than $20\%$ of the difference between baseline RoBERTa-large and RoBERTa-base systems.

We emphasize that our method probes a wide range of speed/accuracy operating points. 
It only requires revisiting task-specific training data,
an expense comparable to fine-tuning,
and does not require revisiting transformer pretraining, a much larger expense comparable to the original pretraining of a transformer model.
Our method is robust across both BERT- and RoBERTa-based models.
It is also robust across both SQuAD and NQ, despite the different biases
incorporated into the construction of these datasets.
Furthermore our observation that the resulting transformer layers are non-identical may inform future efforts at pruning.

\section{Ethical Consideration}
The methods described in this paper are able to reduce the energy-intensiveness of transformer language models, both at runtime, and by reducing the need for pretraining of such models.
All experiments were done with publicly available data sets that are not known to contain personally identifiable information.
Although deployed question answering system have the potential for misuse, this work is not likely to affect this potential.

\bibliography{main}
\bibliographystyle{acl_natbib}

\end{document}